\pdfoutput=1

\documentclass[11pt]{article}

\usepackage[]{naacl2021}

\usepackage{times}
\usepackage{latexsym}

\usepackage{amssymb}
\usepackage{amsmath}
\usepackage{latexsym}
\usepackage{mdwlist}
\usepackage{graphicx}
\usepackage{multirow,booktabs}

\usepackage[T1]{fontenc}

\usepackage[utf8]{inputenc}

\usepackage{microtype}

\title{Semi-Supervised Joint Estimation of Word and Document Readability}

\author{Yoshinari Fujinuma \\
  University of Colorado Boulder \\
  \texttt{fujinumay@gmail.com} \\\And
  Masato Hagiwara \\
  Octanove Labs \\
  \texttt{masato@octanove.com} \\}

\begin{document}
\maketitle
\begin{abstract}
Readability or difficulty estimation of words and documents has been investigated independently in the literature, often assuming the existence of extensive annotated resources for the other. Motivated by our analysis showing that there is a recursive relationship between word and document difficulty, we propose to jointly estimate word and document difficulty through a graph convolutional network (GCN) in a semi-supervised fashion. Our experimental results reveal that the GCN-based method can achieve higher accuracy than strong baselines, and stays robust even with a smaller amount of labeled data.\footnote{Our code is at \url{https://github.com/akkikiki/diff_joint_estimate}}
\end{abstract}

\section{Introduction}

Accurately estimating the readability or difficulty
of words and text has been an important fundamental task in NLP and education, with a wide range of applications including reading resource suggestion~\citep{heilman2008retrieval}, text simplification~\citep{yimam2018cwi}, and automated essay scoring~\citep{vajjala2018experiments}.

A number of linguistic resources have been created either manually or semi-automatically for non-native learners of languages such as English~\citep{capel2010a1,capel2012completing}, French~\cite{francois2014flelex}, and Swedish~\citep{francois2016svalex,alfter2018towards}, often referencing the Common European Framework of Reference~\citep[CEFR]{council2001}. However, few linguistic resources exist outside these major European languages and manually constructing such resources demands linguistic expertise and efforts.

This led to the proliferation of NLP-based {\it readability} or {\it difficulty} {\it assessment} methods to automatically estimate the difficulty of words and texts~\citep{vajjala2012improving,wang2016grammatical,alfter2018towards,vajjala2018experiments,settles2020machine}. However, bootstrapping lexical resources with difficulty information often assumes the existence of textual datasets (e.g., digitized coursebooks) annotated with difficulty. Similarly, many text readability estimation methods \citep{wang2016grammatical,xia2016cambridge} assume the existence of abundant lexical or grammatical resources annotated with difficulty information. Individual research studies focus only on one side, either words or texts, although in reality they are closely intertwined---there is a {\it recursive relationship between word and text difficulty}, where the difficulty of a word is correlated to the {\it minimum} difficulty of the document where that word appears, and the difficulty of a document is correlated to the {\it maximum} difficulty of a word in that document (Figure~\ref{fig:heatmap}).

\begin{figure}[!t]
\begin{center}
\includegraphics[scale=0.3]{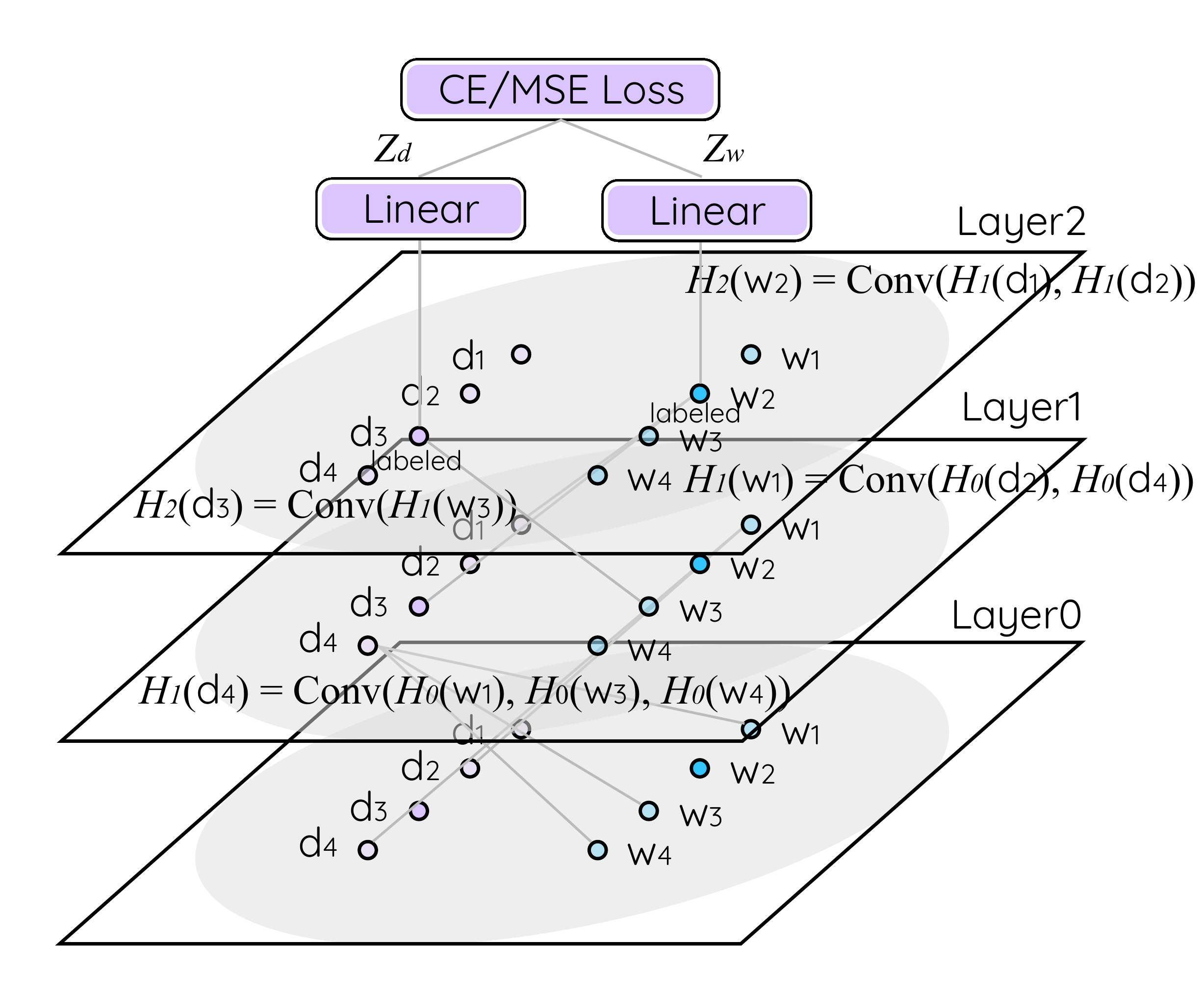} 
\caption{Overview of the proposed GCN architecture which recursively connects word $w_i$ and document $d_j$ to exploit the recursive relationship of their difficulty.}
\label{fig:gcn}
\vspace{-16pt}
\end{center}
\end{figure}

We propose a method to jointly estimate word and text readability in a semi-supervised fashion from a smaller number of labeled data by leveraging the recursive relationship between words and documents. Specifically, we leverage recent developments in graph convolutional networks~\citep[GCNs]{kipf2017semi} and predict the difficulty of words and documents simultaneously by modeling those as nodes in a graph structure and recursively inferring their embeddings using the convolutional layers (Figure~\ref{fig:gcn}). Our model leverages not only the supervision signals but also the recursive nature of word-document relationship. The contributions of this paper are two fold:

\begin{itemize*}
    \item We reframe the word and document readability estimation task as a semi-supervised, joint estimation problem motivated by their recursive relationship of difficulty.
    \item We show that GCNs are effective for solving this by exploiting unlabeled data effectively, even when less labeled data is available.
\end{itemize*}

\section{Task Definition}
\label{sec:task}


Given a set of words $\mathcal{W}$ and documents $\mathcal{D}$, the goal of the joint readability estimation task is to find a function $f$ that maps both words and documents to their difficulty label $f: \mathcal{W} \cup \mathcal{D} \to Y$. Documents here can be text of an arbitrary length, although we use paragraphs as the basic unit of prediction.
This task can be solved as a classification problem or a regression problem where $Y \in \mathbb{R}$.
We use six CEFR-labels representing six levels of difficulty, such as $Y \in \{\text{A1 (lowest), A2, B1, B2, C1, C2 (highest)}\}$ for classification, and a real-valued readability estimate $\beta \in \mathbb{R}$ inspired by the item response theory \citep[IRT]{lord1980irt} for regression\footnote{We assumed the difficulty estimate $\beta$ is normally distributed and used the mid-point of six equal portions of $N(0, 1)$ when mapping CEFR levels to $\beta$.}.
The $\beta$ for each six CEFR level are A1$=-1.38$, A2$=-0.67$,  B1$=-0.21$, B2$=0.21$, C1$=0.67$, and C2$=1.38$.


Words and documents consist of mutually exclusive unlabeled subsets $\mathcal{W}_U$ and $\mathcal{D}_U$ and labeled subsets $\mathcal{W}_L$ and $\mathcal{D}_L$. The function $f$ is inferred using the supervision signal from $\mathcal{W}_L$ and $\mathcal{D}_L$, and potentially other signals from $\mathcal{W}_U$ and $\mathcal{D}_U$ (e.g., relationship between words and documents).

\section{Exploiting Recursive Relationship by Graph Convolutional Networks}

We first show how the readability of words and documents are recursively related to each other.
We then introduce a method based on graph convolutional networks (GCN) to capture such relationship.

\subsection{Recursive Relationship of Word and Document Difficulty}
\begin{figure}[!t]
\begin{center}
\includegraphics[scale=0.4]{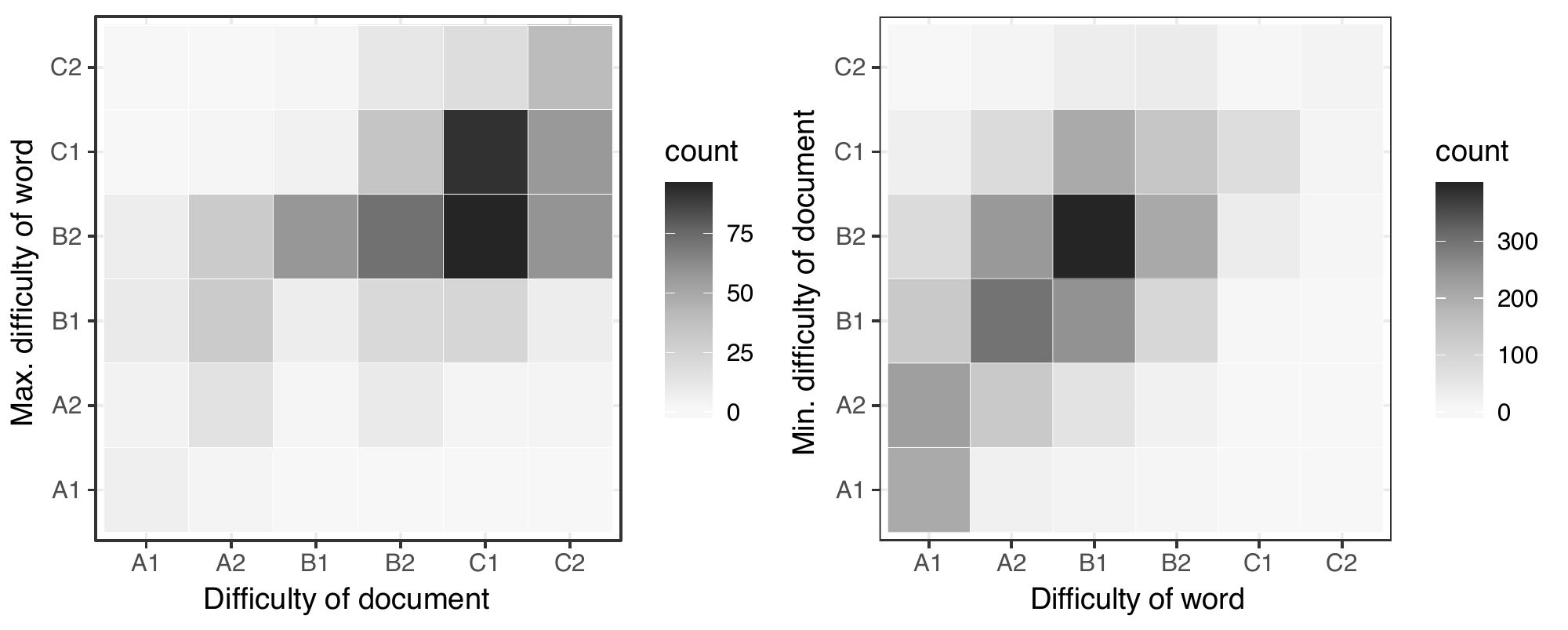} 
\caption{
Recursive relationship of word/document difficulty.
Word difficulty is correlated to the {\it minimum} difficulty of the document where that word appears, and document difficulty is correlated to the {\it maximum} difficulty of a word in that document.
}
\label{fig:heatmap}
\end{center}
\end{figure}

The motivation of using a graph-based method for difficulty classification is the recursive relationship of word and document difficulty.
Figure~\ref{fig:heatmap} shows such recursive relationship using the difficulty-labeled datasets explained in Section 5.
One insight here is the strong correlation between the difficulty of a document and {\it the maximum difficulty of a word in that document}.
This is intuitive and shares motivation with a method which exploits hierarchical structure of a document~\citep{yang2016hierarchical}.
However, the key insight here is the strong correlation between the difficulty of a word and {\it the minimum difficulty of a document where that word appears}, indicating that the readability of words informs that of documents, and vise versa.

%

\subsection{Graph Convolutional Networks on Word-Document Graph}
To capture the recursive, potentially nonlinear relationship between word and document readability while leveraging supervision signals and features, we propose to use graph convolutional networks~\citep[GCNs]{kipf2017semi} specifically built for text classification~\citep{yao2019GCN}, which treats words and documents as nodes.
Intuitively, the hidden layers in GCN, which recursively connects word and document nodes,  encourage exploiting the recursive word-document relationship.

Given a heterogeneous word-document graph $G = (V, E)$ and its adjacency matrix $A \in \mathbb{R}^{|V| \times |V|}$, the hidden states for each layer $H_n \in \mathbb{R}^{|V| \times h_n}$ in a GCN with $N$ hidden layers is computed using the previous layer $H_{n-1}$ as:
\begin{equation}
  H_n = \sigma(\tilde{A} H_{n-1} W_n)
\end{equation}
where $\sigma$ is the ReLU function\footnote{A simplified version of GCN with linear layers \cite{wu2019simplifying} in preliminary experiments shows that hidden layers with ReLU performed better.},
$\tilde{A} = D^{-\frac{1}{2}} A D^{-\frac{1}{2}}$ i.e., a symmetrically normalized matrix of $A$ with its degree matrix $D$, and $W_n \in \mathbb{R}^{h_{n-1} \times h_n}$ is the weight matrix for the $n$th layer.
The input to the first layer $H_1$ is $H_0 = X$ where $X \in \mathbb{R}^{|V| \times h_0}$ is the feature matrix with $h_0$ dimensions for each node in $V$.
We use three different edge weights following~\citet{yao2019GCN}: (1) $A_{ij} = \mbox{tfidf}_{ij}$ if $i$ is a document and $j$ is a word, (2) the normalized point-wise mutual information (PMI) i.e., $A_{ij} = \mbox{PMI}(i, j)$ if both $i$ and $j$ are words, and (3) self-loops, i.e., $A_{ii} = 1$ for all $i$.

We now describe the components which differs from ~\citet{yao2019GCN}. We use separate final linear layers for words and documents\footnote{A model variant with a common linear layer (i.e., original GCN) for both words and documents did not perform as well.}:
\begin{eqnarray}
  Z_w &=& H_{N} W_w + b_w \\
  Z_d &=& H_{N} W_d + b_d
\end{eqnarray}
where $W$ and $b$ are the weight and bias of the layer, and used a linear combination of word and document losses weighted by $\alpha$ (Figure~\ref{fig:gcn})
\begin{equation}
  \mathcal{L} = \alpha \mathcal{L}(Z_w) + (1 - \alpha) \mathcal{L}(Z_d)
\end{equation}

For regression, we used $Z$ ($Z_w$ for words and $Z_d$ for documents) as the prediction of node $v$ and used the mean squared error (MSE):
\begin{equation}
   \mathcal{L}(Z) = \frac{1}{|V_L|} \sum_{v \in V_L} (Z_v - Y_v)^2
\end{equation}
where $V_L = \mathcal{W}_L \cup \mathcal{D}_L$ is the set of labeled nodes. For classification, we use a softmax layer followed by a cross-entropy (CE) loss:
\begin{equation}
    \mathcal{L}(Z) = -\sum_{v \in V_L} \log \frac{\exp(Z_{v, Y_v})}{\sum_i \exp(Z_{v, i})}.
\end{equation}
Since GCN is transductive, node set $V$ also includes the unlabeled nodes from the evaluation sets and have predicted difficulty labels assigned when training is finished.

\section{Experiments}

\paragraph{Datasets}
We use publicly available English CEFR-annotated resources for second language learners, such as CEFR-J~\citep{negishi2013progress} Vocabulary Profile as words and Cambridge English Readability Dataset~\citep{xia2016cambridge} as documents (Table~\ref{tab:dataset}). Since these two datasets lack C1/C2-level words and A1 documents, we hired a linguistic PhD to write these missing portions\footnote{The dataset is available at \url{https://github.com/openlanguageprofiles/olp-en-cefrj}.}.

\paragraph{Baselines}
We compare our method against methods used in previous work~\citep{feng2010readability,vajjala2012improving,martinc2019readability,deutsch2020linguistic}: (1) logistic regression for classification (LR cls),
(2) linear regression for regression (LR regr),
(3) Gradient Boosted Decision Tree (GBDT), and
(4) Hierarchical Attention Network \citep[HAN]{yang2016hierarchical}, which is reported as one of the state-of-the-art methods in readability assessment for documents \citep{martinc2019readability,deutsch2020linguistic}.

\paragraph{Features}
\label{sec:features}
For all methods except for HAN, we use both surface or ``traditional''~\citep{vajjala2012improving} and embedding features on words and documents which are shown to be effective for readability estimation ~\citep{culligan2015comparison,settles2020machine,deutsch2020linguistic}.
For words, we use their length (in characters), the log frequency in Wikipedia \citep{filip2017wikidump}, 
and GloVe~\citep{pennington2014glove}.
For documents, we use the number of NLTK~\citep{loper2002nltk}-tokenized words in a document, and the output of embeddings from BERT-base model~\citep{devlin2019bert} which are averaged over all tokens in a given sentence.

\begin{table}[]
\centering
{\small
\begin{tabular}{lrrr}
\toprule
Dataset                      & Train  & Dev   & Test \\ \midrule
Words (CEFR-J + C1/C2)       & 2,043  & 447   & 389  \\
Documents (Cambridge + A1)   &   482  &  103  &  98  \\ \bottomrule
\end{tabular}
}
 \caption{Dataset size for words and documents}
 \label{tab:dataset}
\end{table}

\paragraph{Hyperparameters}
We conduct random hyperparameter search with $200$ samples,
separately selecting two different sets of hyperparameters, one optimized for word difficulty and the other for document.
We set the number of hidden layers $N = 2$ with $h_n = 512$ for documents and $N = 1$ with $h_n = 64$ for words.
See Appendix A for the details on other hyperparameters.

\paragraph{Evaluation}
We use accuracy and Spearman's rank correlation as the metrics.
When calculating the correlation for a classification model, we convert the discrete outputs into continuous values in two ways: (1) convert the CEFR label with the maximum probability into corresponding $\beta$ in Section~\ref{sec:task}, (cls+m), or (2) take a sum of all $\beta$ in six labels weighted by their probabilities (cls+w).

\begin{table}
\centering
\begin{tabular}{lrrrr}
\toprule
                     & \multicolumn{2}{c}{Word} & \multicolumn{2}{c}{Document}        \\
   Method           &  Acc      &   Corr &  Acc       & Corr     \\ \midrule
HAN            &  -             & -         &  0.367 & 0.498                     \\
LR (regr)      & 0.409         & 0.534    & 0.480          &  0.657             \\
LR (cls+m)     &  0.440        & 0.514    & 0.765          & 0.723              \\
LR (cls+w)     &  0.440        & 0.540    & 0.765          & 0.880              \\
GBDT           & 0.432         & 0.376         &   0.765            &  0.833                  \\
GCN (regr)           &  0.434       & 0.579    & 0.643        & 0.849             \\
GCN (cls+m)          &  \bf 0.476       & 0.536    & \bf 0.796          & 0.878              \\
GCN (cls+w)          &  \bf 0.476       & \bf 0.592    & \bf 0.796          & \bf 0.891              \\ \bottomrule
\end{tabular}
\caption{
Difficulty estimation results in accuracy (Acc) and correlation (Corr) on classification outputs converted to continuous values by taking the max (cls+m) or weighted sum (cls+w) and regression (regr) variants for the logistic regression (LR) and GCN.
}
\label{tab:without_unlabeled}
\end{table}

\subsection{Results}
\label{sec:labled_results}
Table~\ref{tab:without_unlabeled} shows the test accuracy and correlation results.
GCNs show increase in both document accuracy and word accuracy compared to the baseline. We infer that this is because GCN is good at capturing the relationship between words and documents.
For example, the labeled training documents include an A1 document and that contains the word ``bicycle,'' and the difficulty label of the document is explicitly propagated to the ``bicycle'' word node, whereas the logistic regression baseline mistakenly predicts as A2-level, since it relies solely on the input features to capture its similarities.


\subsection{Ablation Study on Features}
Table~\ref{tab:ablation} shows the ablation study on the features explained in Section~\ref{sec:features}.
By comparing Table~\ref{tab:without_unlabeled}  and Table~\ref{tab:ablation}, which are experimented on the same datasets, GCN without using any traditional or embedding features (``None'') shows comparative results to some baselines, especially on word-level accuracy.
Therefore, the structure of the word-document graph provides effective and complementary signal for readability estimation.

Overall, the BERT embedding is a powerful feature for predicting document readability on Cambridge Readabilty Dataset. Ablating the BERT embeddings (Table~\ref{tab:ablation}) significantly decreases the document accuracy ($-0.112$) which is consistent with the previous work~\citep{martinc2019readability,deutsch2020linguistic} that BERT being one of the best-performing method for predicting document readability on one of the datasets they used, and HAN performing relatively low due to not using the BERT embeddings. 

\begin{table}
\centering
\begin{tabular}{lrrrr}
\toprule
             & \multicolumn{2}{c}{Word} & \multicolumn{2}{c}{Document}        \\
Features     & Acc    & Corr  & Acc   & Corr     \\ \midrule
All          & 0.476  & 0.592 & 0.796 & 0.891 \\
$-$word freq.& 0.476  & 0.591 & 0.796 & 0.899 \\
$-$doc length& 0.481  & 0.601 & 0.796 & 0.890 \\
$-$GloVe     & 0.463  & 0.545 & 0.714 & 0.878 \\
$-$BERT      & 0.450  & 0.547 & 0.684 & 0.830  \\
None         & 0.440  & 0.436 & 0.520 & 0.669 \\
\bottomrule
\end{tabular}
 \caption{
 Ablation study on the features used.
 ``None'' is when applying GCN without any features ($X=I$ i.e., one-hot encoding per node), which solely relies on the word-document structure of the graph.
 }
 \label{tab:ablation}
\end{table}

\subsection{Training on Less Labeled Data}

\begin{figure}[!t]
\begin{center}
\includegraphics[width=.95\linewidth]{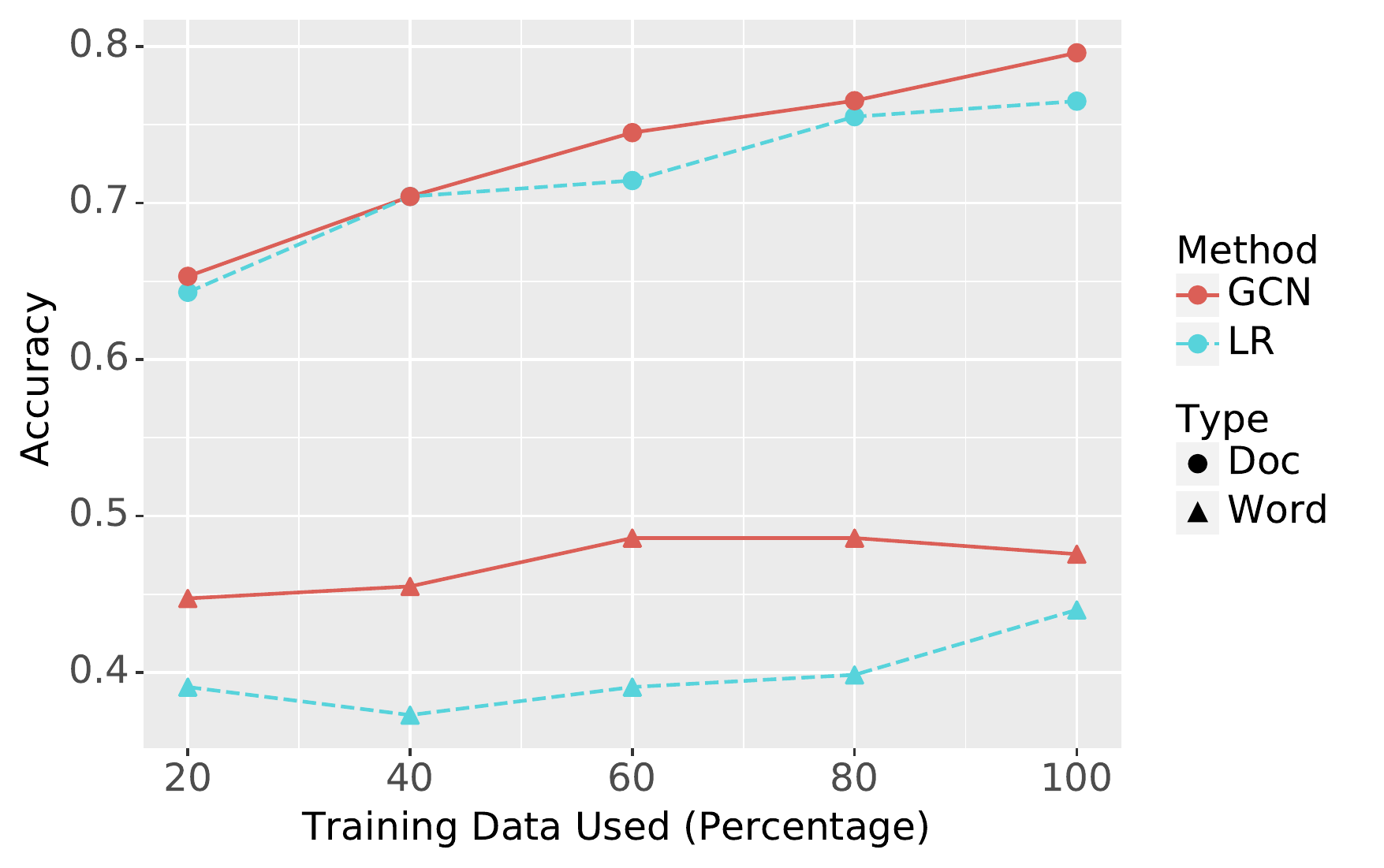} 
\caption{
Word and document accuracy with different amount of training data used.}
\label{fig:train_portion}
\end{center}
\end{figure}

To analyze whether GCN is robust when training dataset is small, we compare the baseline and GCN by varying the amount of labeled training data. In Figure~\ref{fig:train_portion}, we observe consistent improvement in GCN over the baseline especially in word accuracy. This outcome suggests that the performance of GCN stays robust even with smaller training data by exploiting the signals gained from the recursive word-document relationship and their structure.
Another trend observed in Figure~\ref{fig:train_portion} is the larger gap in word accuracy compared to document accuracy when the training data is small likely due to GCN explicitly using context given by word-document edges.

\section{Conclusion}

In this paper, we proposed a GCN-based method to jointly estimate the readability on both words and documents. We experimentally showed that GCN achieves higher accuracy by capturing the recursive difficulty relationship between words and documents, even when using a smaller amount of labeled data. GCNs are a versatile framework that allows inclusion of diverse types of nodes, such as subwords, paragraphs, and even grammatical concepts. We leave this investigation as future work.

\section*{Acknowledgements}
The authors would like to thank Adam Wiemerslage, Michael J. Paul, and anonymous reviewers for their detailed and constructive feedback. We also thank Kathleen Hall for her help with annotation.

\bibliography{custom}

\begin{thebibliography}{27}
\expandafter\ifx\csname natexlab\endcsname\relax\def\natexlab#1{#1}\fi

\bibitem[{Alfter and Volodina(2018)}]{alfter2018towards}
David Alfter and Elena Volodina. 2018.
\newblock \href {https://doi.org/10.18653/v1/W18-0508} {Towards single word
  lexical complexity prediction}.
\newblock In \emph{Proceedings of the Thirteenth Workshop on Innovative Use of
  {NLP} for Building Educational Applications}.

\bibitem[{Capel(2010)}]{capel2010a1}
Annette Capel. 2010.
\newblock {A1}--{B2} vocabulary: insights and issues arising from the {E}nglish
  {P}rofile {W}ordlists project.
\newblock \emph{English Profile Journal}, 1.

\bibitem[{Capel(2012)}]{capel2012completing}
Annette Capel. 2012.
\newblock Completing the english vocabulary profile: {C1} and {C2} vocabulary.
\newblock \emph{English Profile Journal}, 3.

\bibitem[{{Council of Europe}(2001)}]{council2001}
{Council of Europe}. 2001.
\newblock \emph{Common European Framework of Reference for Languages: Learning,
  Teaching, Assessment}.
\newblock Press Syndicate of the University of Cambridge.

\bibitem[{Culligan(2015)}]{culligan2015comparison}
Brent Culligan. 2015.
\newblock A comparison of three test formats to assess word difficulty.
\newblock \emph{Language Testing}, 32(4):503--520.

\bibitem[{Deutsch et~al.(2020)Deutsch, Jasbi, and
  Shieber}]{deutsch2020linguistic}
Tovly Deutsch, Masoud Jasbi, and Stuart Shieber. 2020.
\newblock \href {https://doi.org/10.18653/v1/2020.bea-1.1} {Linguistic features
  for readability assessment}.
\newblock In \emph{Proceedings of the Fifteenth Workshop on Innovative Use of
  NLP for Building Educational Applications}.

\bibitem[{Devlin et~al.(2019)Devlin, Chang, Lee, and
  Toutanova}]{devlin2019bert}
Jacob Devlin, Ming-Wei Chang, Kenton Lee, and Kristina Toutanova. 2019.
\newblock \href {https://doi.org/10.18653/v1/N19-1423} {{BERT}: {P}re-training
  of deep bidirectional transformers for language understanding}.
\newblock In \emph{Proceedings of the North American Chapter of the Association
  for Computational Linguistics: Human Language Technologies}.

\bibitem[{Feng et~al.(2010)Feng, Jansche, Huenerfauth, and
  Elhadad}]{feng2010readability}
Lijun Feng, Martin Jansche, Matt Huenerfauth, and No^^c3^^a9mie Elhadad. 2010.
\newblock \href {http://www.aclweb.org/anthology/C10-2032} {A comparison of
  features for automatic readability assessment}.

\bibitem[{Fran{\c{c}}ois et~al.(2014)Fran{\c{c}}ois, Gala, Watrin, and
  Fairon}]{francois2014flelex}
Thomas Fran{\c{c}}ois, N{\`u}ria Gala, Patrick Watrin, and C{\'e}drick Fairon.
  2014.
\newblock \href
  {http://www.lrec-conf.org/proceedings/lrec2014/pdf/1108_Paper.pdf} {{FLEL}ex:
  a graded lexical resource for {F}rench foreign learners}.
\newblock In \emph{Proceedings of the Language Resources and Evaluation
  Conference}.

\bibitem[{Fran{\c{c}}ois et~al.(2016)Fran{\c{c}}ois, Volodina, Pil{\'a}n, and
  Tack}]{francois2016svalex}
Thomas Fran{\c{c}}ois, Elena Volodina, Ildik{\'o} Pil{\'a}n, and Ana{\"\i}s
  Tack. 2016.
\newblock \href {https://www.aclweb.org/anthology/L16-1032} {{SVAL}ex: a
  {CEFR}-graded lexical resource for {S}wedish foreign and second language
  learners}.
\newblock In \emph{Proceedings of the Language Resources and Evaluation
  Conference}.

\bibitem[{Ginter et~al.(2017)Ginter, Haji{\v c}, Luotolahti, Straka, and
  Zeman}]{filip2017wikidump}
Filip Ginter, Jan Haji{\v c}, Juhani Luotolahti, Milan Straka, and Daniel
  Zeman. 2017.
\newblock \href {http://hdl.handle.net/11234/1-1989} {{CoNLL} 2017 shared task
  - automatically annotated raw texts and word embeddings}.
\newblock {LINDAT}/{CLARIAH}-{CZ} digital library at the Institute of Formal
  and Applied Linguistics ({{\'U}FAL}), Faculty of Mathematics and Physics,
  Charles University.

\bibitem[{Heilman et~al.(2008)Heilman, Zhao, Pino, and
  Eskenazi}]{heilman2008retrieval}
Michael Heilman, Le~Zhao, Juan Pino, and Maxine Eskenazi. 2008.
\newblock \href {https://www.aclweb.org/anthology/W08-0910} {Retrieval of
  reading materials for vocabulary and reading practice}.
\newblock In \emph{Proceedings of the Third Workshop on Innovative Use of {NLP}
  for Building Educational Applications}, pages 80--88.

\bibitem[{Kipf and Welling(2017)}]{kipf2017semi}
Thomas~N. Kipf and Max Welling. 2017.
\newblock Semi-supervised classification with graph convolutional networks.
\newblock In \emph{Proceedings of the International Conference on Learning
  Representations}.

\bibitem[{Loper and Bird(2002)}]{loper2002nltk}
Edward Loper and Steven Bird. 2002.
\newblock {NLTK}: {T}he natural language toolkit.
\newblock In \emph{Proceedings of the ACL Workshop on Effective Tools and
  Methodologies for Teaching Natural Language Processing and Computational
  Linguistics}.

\bibitem[{Lord(1980)}]{lord1980irt}
Frederic~M. Lord. 1980.
\newblock \emph{Applications of Item Response Theory To Practical Testing
  Problems}.
\newblock Lawrence Erlbaum Associates.

\bibitem[{Martinc et~al.(2019)Martinc, Pollak, and
  Robnik{-}Sikonja}]{martinc2019readability}
Matej Martinc, Senja Pollak, and Marko Robnik{-}Sikonja. 2019.
\newblock \href {http://arxiv.org/abs/1907.11779} {Supervised and unsupervised
  neural approaches to text readability}.
\newblock \emph{CoRR}, abs/1907.11779.

\bibitem[{Negishi et~al.(2013)Negishi, Takada, and Tono}]{negishi2013progress}
Masashi Negishi, Tomoko Takada, and Yukio Tono. 2013.
\newblock A progress report on the development of the {CEFR-J}.
\newblock In \emph{Exploring language frameworks: {P}roceedings of the ALTE
  Krak{\'o}w Conference}, pages 135--163.

\bibitem[{Pennington et~al.(2014)Pennington, Socher, and
  Manning}]{pennington2014glove}
Jeffrey Pennington, Richard Socher, and Christopher~D. Manning. 2014.
\newblock \href {https://doi.org/10.3115/v1/D14-1162} {{GloVe}: {G}lobal
  vectors for word representation}.
\newblock In \emph{Proceedings of Empirical Methods in Natural Language
  Processing}.

\bibitem[{Settles et~al.(2020)Settles, T.~LaFlair, and
  Hagiwara}]{settles2020machine}
Burr Settles, Geoffrey T.~LaFlair, and Masato Hagiwara. 2020.
\newblock \href {https://doi.org/10.1162/tacl\_a\_00310} {Machine
  learning^^e2^^80^^93driven language assessment}.
\newblock \emph{Transactions of the Association for Computational Linguistics},
  8:247--263.

\bibitem[{Vajjala and Meurers(2012)}]{vajjala2012improving}
Sowmya Vajjala and Detmar Meurers. 2012.
\newblock \href {https://www.aclweb.org/anthology/W12-2019} {On improving the
  accuracy of readability classification using insights from second language
  acquisition}.
\newblock In \emph{Proceedings of the Seventh Workshop on Building Educational
  Applications Using {NLP}}.

\bibitem[{Vajjala and Rama(2018)}]{vajjala2018experiments}
Sowmya Vajjala and Taraka Rama. 2018.
\newblock \href {https://doi.org/10.18653/v1/W18-0515} {Experiments with
  universal {CEFR} classification}.
\newblock In \emph{Proceedings of the Thirteenth Workshop on Innovative Use of
  {NLP} for Building Educational Applications}.

\bibitem[{Wang and Andersen(2016)}]{wang2016grammatical}
Shuhan Wang and Erik Andersen. 2016.
\newblock \href {https://www.aclweb.org/anthology/C16-1159} {Grammatical
  templates: Improving text difficulty evaluation for language learners}.

\bibitem[{Wu et~al.(2019)Wu, Zhang, de~Souza~Jr., Fifty, Yu, and
  Weinberger}]{wu2019simplifying}
Felix Wu, Tianyi Zhang, Amauri~Holanda de~Souza~Jr., Christopher Fifty, Tao Yu,
  and Kilian~Q. Weinberger. 2019.
\newblock Simplifying graph convolutional networks.
\newblock In \emph{Proceedings of the International Conference of Machine
  Learning}.

\bibitem[{Xia et~al.(2016)Xia, Kochmar, and Briscoe}]{xia2016cambridge}
Menglin Xia, Ekaterina Kochmar, and Ted Briscoe. 2016.
\newblock \href {https://doi.org/10.18653/v1/W16-0502} {Text readability
  assessment for second language learners}.
\newblock In \emph{Proceedings of the 11th Workshop on Innovative Use of {NLP}
  for Building Educational Applications}.

\bibitem[{Yang et~al.(2016)Yang, Yang, Dyer, He, Smola, and
  Hovy}]{yang2016hierarchical}
Zichao Yang, Diyi Yang, Chris Dyer, Xiaodong He, Alex Smola, and Eduard Hovy.
  2016.
\newblock \href {https://doi.org/10.18653/v1/N16-1174} {Hierarchical attention
  networks for document classification}.
\newblock In \emph{Proceedings of the North American Chapter of the Association
  for Computational Linguistics: Human Language Technologies}.

\bibitem[{Yao et~al.(2019)Yao, Mao, and Luo}]{yao2019GCN}
Liang Yao, Chengsheng Mao, and Yuan Luo. 2019.
\newblock Graph convolutional networks for text classification.
\newblock In \emph{Association for the Advancement of Artificial Intelligence}.

\bibitem[{Yimam et~al.(2018)Yimam, Biemann, Malmasi, Paetzold, Specia,
  {\v{S}}tajner, Tack, and Zampieri}]{yimam2018cwi}
Seid~Muhie Yimam, Chris Biemann, Shervin Malmasi, Gustavo Paetzold, Lucia
  Specia, Sanja {\v{S}}tajner, Ana{\"\i}s Tack, and Marcos Zampieri. 2018.
\newblock \href {https://doi.org/10.18653/v1/W18-0507} {A report on the complex
  word identification shared task 2018}.
\newblock In \emph{Proceedings of the Thirteenth Workshop on Innovative Use of
  {NLP} for Building Educational Applications}.

\end{thebibliography}
\bibliographystyle{acl_natbib}

\appendix
\section{Hyperparameter Details}
\label{sec:hyperparameters}
We conduct random hyperparameter search with $200$ samples in the following ranges:
$\alpha \in \{0.1, 0.2, ..., 0.9\}$,
the learning rate from $\{1, 2, 5, 10, 20, 50, 100\}\times 10^{-4}$,
dropout probability from $\{0.1, 0.2, ..., 0.5\}$,
the number of epochs from $\{250, 500, 1000, 1500, 2000\}$,
the number of hidden units $h_n \in \{32, 64, 128, 256, 512, 1024\}$,
the number of hidden layers from $\{1, 2, 3\}$, and the
PMI window width from $\{\mathrm{disabled}, 5, 10, 15, 20\}$.

We now describe the selected best combination of hyperparameters for each setting.
For GCN in the classification setting,
the selected hyperparameters for document difficulty estimation are:
\begin{itemize*}
 \item $\alpha$: $0.3$
 \item Learning rate: $5 \cdot 10^{-4}$
 \item Dropout probability: $0.5$
 \item The number of epochs: $500$
 \item The number of hidden units $h_n$: $512$
 \item The number of hidden layers $N$: $2$
 \item PMI window width: $5$
\end{itemize*}
and for word difficulty estimation, the selected hyperparameters are:
\begin{itemize*}
 \item $\alpha$: $0.2$
 \item Learning rate: $5 \cdot 10^{-3}$
 \item Dropout probability: $0.2$
 \item The number of epochs: $250$
 \item The number of hidden units $h_n$: $64$
 \item The number of hidden layers $N$: $1$
 \item PMI window width: $\mathrm{disabled}$
\end{itemize*}

For GCN in the regression setting,
the selected hyperparameters for document difficulty estimation are:
\begin{itemize*}
 \item $\alpha$: $0.4$
 \item Learning rate: $2 \cdot 10^{-4}$
 \item Dropout probability: $0.3$
 \item The number of epochs: $1500$
 \item The number of hidden units $h_n$: $128$
 \item The number of hidden layers $N$: $2$
 \item PMI window width: $5$
\end{itemize*}
and for word difficulty estimation, the selected hyperparameters are:
\begin{itemize*}
 \item $\alpha$: $0.2$
 \item Learning rate: $1 \cdot 10^{-3}$
 \item Dropout probability: $0.1$
 \item The number of epochs: $500$
 \item The number of hidden units $h_n$: $512$
 \item The number of hidden layers $N$: $2$
 \item PMI window width: $\mathrm{disabled}$
\end{itemize*}
\end{document}